\documentclass[10pt, letter, conference]{IEEEtran}
\IEEEoverridecommandlockouts
\usepackage{cite}
\usepackage{amsbsy,amsmath,amssymb,amsfonts}
\usepackage{mathtools}
\usepackage{commath}

\newcommand\mysim{\mathrel{\stackrel{\makebox[0pt]{\mbox{\normalfont\tiny U}}}{\sim}}}

\newcommand{\argmin}{\mathop{\mathrm{argmin}}}
\newcommand{\RN}[1]{%
  \textup{\uppercase\expandafter{\romannumeral#1}}%
}

\DeclareMathOperator{\E}{\mathbb{E}}

\usepackage[super]{nth}
\usepackage[ruled,vlined]{algorithm2e}

\DeclarePairedDelimiter\floor{\lfloor}{\rfloor}

\usepackage{algorithmic}
\usepackage{graphicx}
\usepackage{textcomp}
\usepackage{xcolor}

\def\BibTeX{{\rm B\kern-.05em{\sc i\kern-.025em b}\kern-.08em
    T\kern-.1667em\lower.7ex\hbox{E}\kern-.125emX}}

\usepackage{tikz}
\newcommand\copyrighttext{
  \footnotesize \copyright 2021 IEEE. Personal use of this material is permitted.  Permission from IEEE must be obtained for all other uses, in any current or future media, including reprinting/republishing this material for advertising or promotional purposes, creating new collective works, for resale or redistribution to servers or lists, or reuse of any copyrighted component of this work in other works.}
\newcommand\copyrightnotice{
\begin{tikzpicture}[remember picture,overlay]
\node[anchor=south,yshift=10pt] at (current page.south) {\fbox{\parbox{\dimexpr\textwidth-\fboxsep-\fboxrule\relax}{\copyrighttext}}};
\end{tikzpicture}
}

\begin{document}
\title{Adversarial Multi-task Learning Enhanced Physics-informed Neural Networks for Solving Partial Differential Equations\\
}
\author{\IEEEauthorblockN{Pongpisit Thanasutives\IEEEauthorrefmark{2}, Masayuki Numao\IEEEauthorrefmark{3}, Ken-ichi Fukui\IEEEauthorrefmark{3}}
\IEEEauthorblockA{\IEEEauthorrefmark{2}Graduate School of Information Science and Technology, Osaka University, Japan\\
\IEEEauthorrefmark{3}The Institute of Scientific and Industrial Research, Osaka University, Japan\\
\{thanasutives, fukui\}@ai.sanken.osaka-u.ac.jp}
}
\maketitle
% For arxiv publishing
\copyrightnotice
\begin{abstract}
Recently, researchers have utilized neural networks to accurately solve partial differential equations (PDEs), enabling the mesh-free method for scientific computation. Unfortunately, the network performance drops when encountering a high nonlinearity domain. To improve the generalizability, we introduce the novel approach of employing multi-task learning techniques, the uncertainty-weighting loss and the gradients surgery, in the context of learning PDE solutions. The multi-task scheme exploits the benefits of learning shared representations, controlled by cross-stitch modules, between multiple related PDEs, which are obtainable by varying the PDE parameterization coefficients, to generalize better on the original PDE. Encouraging the network pay closer attention to the high nonlinearity domain regions that are more challenging to learn, we also propose adversarial training for generating supplementary high-loss samples, similarly distributed to the original training distribution. In the experiments, our proposed methods are found to be effective and reduce the error on the unseen data points as compared to the previous approaches in various PDE examples, including high-dimensional stochastic PDEs.
\end{abstract}

\section{Introduction}
Deep learning has tremendous successes in several subfields of artificial intelligence and scientific computing including computer vision, natural language processing, robotics; moreover, deep learning for physics simulation \cite{carleo2019machine} is an attractive field. As a result of advances in automatic differentiation \cite{baydin2017automatic} for estimating derivatives with respect to any input dimension, lately, neural networks have been extensively studied as an alternative way to numerically solve partial differential equations (PDEs) without reliance on spatial-temporal grids. Solving PDEs using neural networks plays an essential role in the development of a hybrid system between physics and machine learning, owing to the physically consistent network outputs as the PDE solutions.

An influential work by Raissi \textit{el al.} \cite{raissi2019physics}, physics informed neural networks (PINNs), carefully designs the solver network loss to be constrained by the underlying PDE and the boundary conditions. Also in \cite{raissi2018forward}, Raissi has cracked the way to solve coupled forward-backward stochastic differential equations (FBSDEs) based on minimizing the loss of a network structure called FBSNN, which is constructed from Euler-Maruyama discretization. However, in practice, the generalizability problem of deep learning is hard to avoid since the network could not always be trained well in high-dimensional non-convex optimization with limited training data, the sampling distribution might not be effective enough \cite{al2018solving} and the overfitting problem, where the network performs worse in spite of the greater performance in the training set. This motivates us to let multi-task learning and adversarial training come to the rescue. To the best of our knowledge, we are the first to leverage this combination of learning techniques in the context of solving PDEs.

The construction of the multi-task setting is natural as most PDEs are parameterized by certain coefficients. We propose that the varied coefficient can be viewed as an auxiliary task that shares additional insights to the original task. Multi-task learning could be implemented in both network architecture and optimization process. We presume that the following conditions hold true for our method to be effective: (1) The neural network is capable of exploiting the knowledge of discrepancy, which is induced by equation's coefficient modification, between multiple related PDEs to achieve the generalized performance. (2) Minimizing the designed physics-informed loss function converges the predicted solution to the true PDE solution. As long as (2), which is theoretically supported by the Theorem 1. in \cite{van2020optimally}, holds, we could apply the novel multi-task learning strategies without supervision.

In this paper, the cross-stitch module \cite{misra2016cross} is selected as the component for learning how to share the information between tasks. The module works well in the case of 2-task learning, in which we are interested. To optimize the multi-task architecture, we experiment with (1) the uncertainty-weighting loss \cite{kendall2018multi} since the strategy needs no extra hyperparameter tuning and potentially excel at training with low-noise or noiseless tasks such as solving for the solutions that fit the PDE constraints. Alternatively, we also try (2) the gradients surgery algorithm (PCGrad) \cite{DBLP:conf/nips/YuK0LHF20} to the unweighted losses to enhance the conformity of gradients magnitude and direction between tasks. We provide a qualitative comparison of these strategies for each PDE problem in Section \ref{results}.

Moreover, the sampling distribution for training data generation is also of concern. Particularly, in the case of solving forward PDE where the sampling process is performed to construct data points in the spatial-temporal grid before training. Previous work \cite{al2018solving} has employed naive uniform sampling and inspected performance degradation on the less emphasized regions. Similar observations are found in \cite{van2020optimally} that the original PINNs perform suboptimally on the high-frequency domain regions where greater attention is needed in order for the network to fit well. As visualized in Fig. \ref{fig:transformed_samples}(A and B), we tackle the problem by including more difficult samples (red points) through our proposed adversarial training while maintaining the characteristics of the training distribution. This is valid since the computational complexity does not blow up rapidly as the number of training points increases \cite{xu2020solving}. 

Inspired by the success of generative adversarial networks (GAN) \cite{goodfellow2014generative}, the difficult samples are defined so that they maximize the loss of the solver network, accounting for the lack of samples in high-loss regions. We also control the generated samples to be similarly distributed to the training distribution. This helps prevent the solver network from being laser-focused or overfitting a particular domain. In our forward PDE experiments (See Section \ref{FPDE}), we apply adversarial training to the best variation of multi-task learning optimization techniques to obtain more superior results, which are also reported in Section \ref{results}.

\section{Related Works}
\subsection{Neural networks for solving PDEs}
Early researchers studied the possibility of employing a universal approximator \cite{csaji2001approximation} based on a neural network to represent PDE solutions back in \cite{dissanayake1994neural, lagaris1998artificial}, resulting in an ability to obtain approximated solutions fast. However, those methods rely on the supervision of referenced solutions, which are achievable analytically or numerically. This is either not practical or limiting the performance of the network since the exact solutions are not always available and numerical approximations do not precisely represent the solutions.

\begin{figure}
  \centerline{\includegraphics[width=0.5\textwidth]{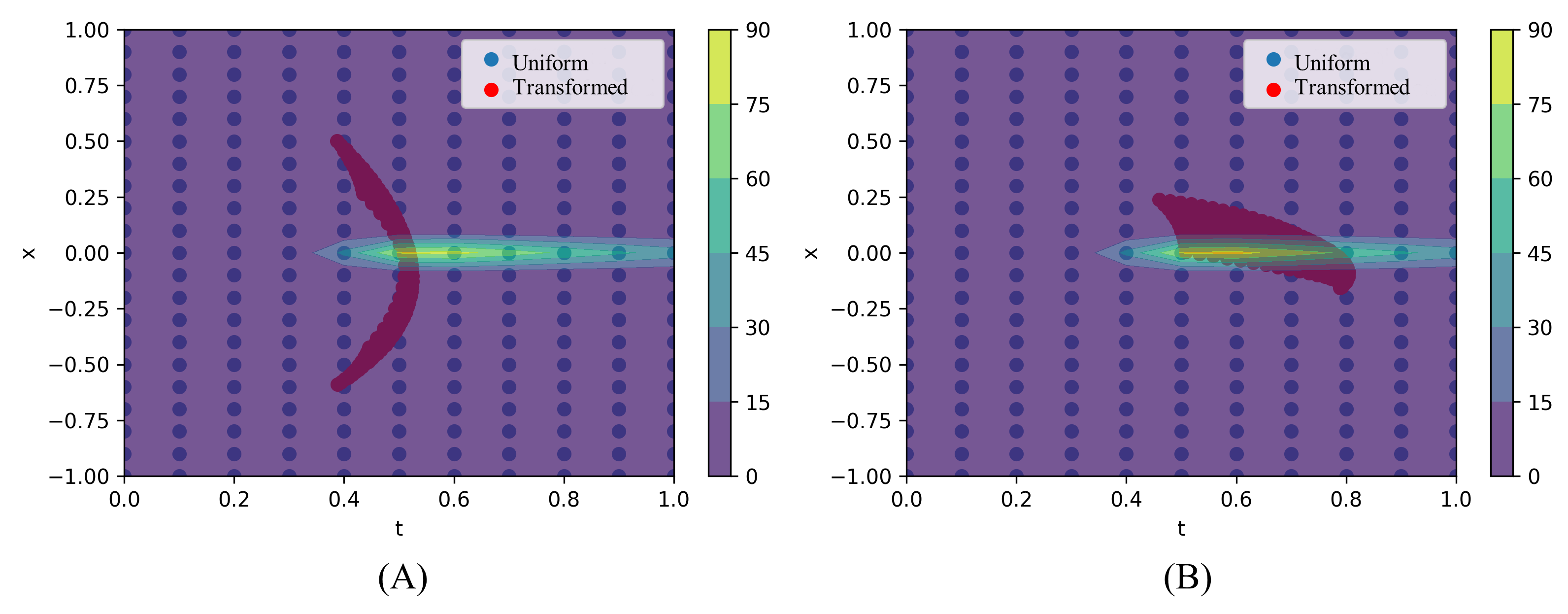}}
  \caption{In the Burgers' equation example, we use uniform distribution to sample training data points and calculate the corresponding losses. (A) and (B) are examples of the transformed samples (from different iterations) in the process of our proposed adversarial training. The color bar indicates the trained PINN loss magnitude. The transformed samples overlap the regions, where more samples are required, and inherit the uniform patterns.}
  \label{fig:transformed_samples}
  \vspace{-4mm}
\end{figure}

Nowadays, PDE solution's derivatives with respect to any independent variables could be easily estimated using automatic differentiation on a computational graph, emerging the rise of neural network applications for solving several kinds of PDE such as \cite{raissi2018forward, han2018solving, sirignano2018dgm, raissi2019physics, xu2020solving} to name but a few. Raissi \textit{et al}. \cite{raissi2019physics} approximated the first and higher-order derivatives using automatic differentiation, whilst Sirignano \textit{et al}. \cite{sirignano2018dgm} leveraged the Monte Carlo method for higher-order derivatives approximation, proposing a so-called Deep Galerkin Method (DGM). Both of them layout the similar fundamentals of considering PDE constraints along with the initial and boundary conditions as the neural network loss function.

Due to the power of the neural network for overcoming the ``curse of dimensionality" in practice, more previous works, FBSNN \cite{raissi2018forward} and Deep BSDE \cite{han2018solving}, extended the framework to solve a large class of high-dimensional nonlinear PDEs by taking a neural network to estimate intractable mathematical expressions from each step of the multi-step prediction process, which was constructed by Euler–Maruyama discretization. Nonetheless, as mentioned, drawbacks of neural networks as PDE solvers were spotted in the past literature.

Van der Meer \textit{et al}. \cite{van2020optimally} found that typical single-objective optimization without scaling factors for balancing between the losses on PDE constraints and boundary conditions could deteriorate the overall performance of the network; therefore, a heuristic method to approximate the optimal scaling parameter, with respect to the relative error, was proposed.

Al-Aradi \textit{et al}. \cite{al2018solving} pointed out that the data sampling method could be one possible source of the model's generalization error. Specifically for the high-dimensional PDEs, Güler \textit{et al}. \cite{guler2019towards} tackled the generalization problem and improved training stability via the network architecture modification (e.g. more residual connections). Differently from \cite{guler2019towards}, we focus more on how multi-task and adversarial training, which are substantially developed in the context of data-driven learning, could be used to alleviate the generalization problem and enable an effective training strategy for solving PDEs.

\begin{figure}
  \centerline{\includegraphics[width=0.4\textwidth]{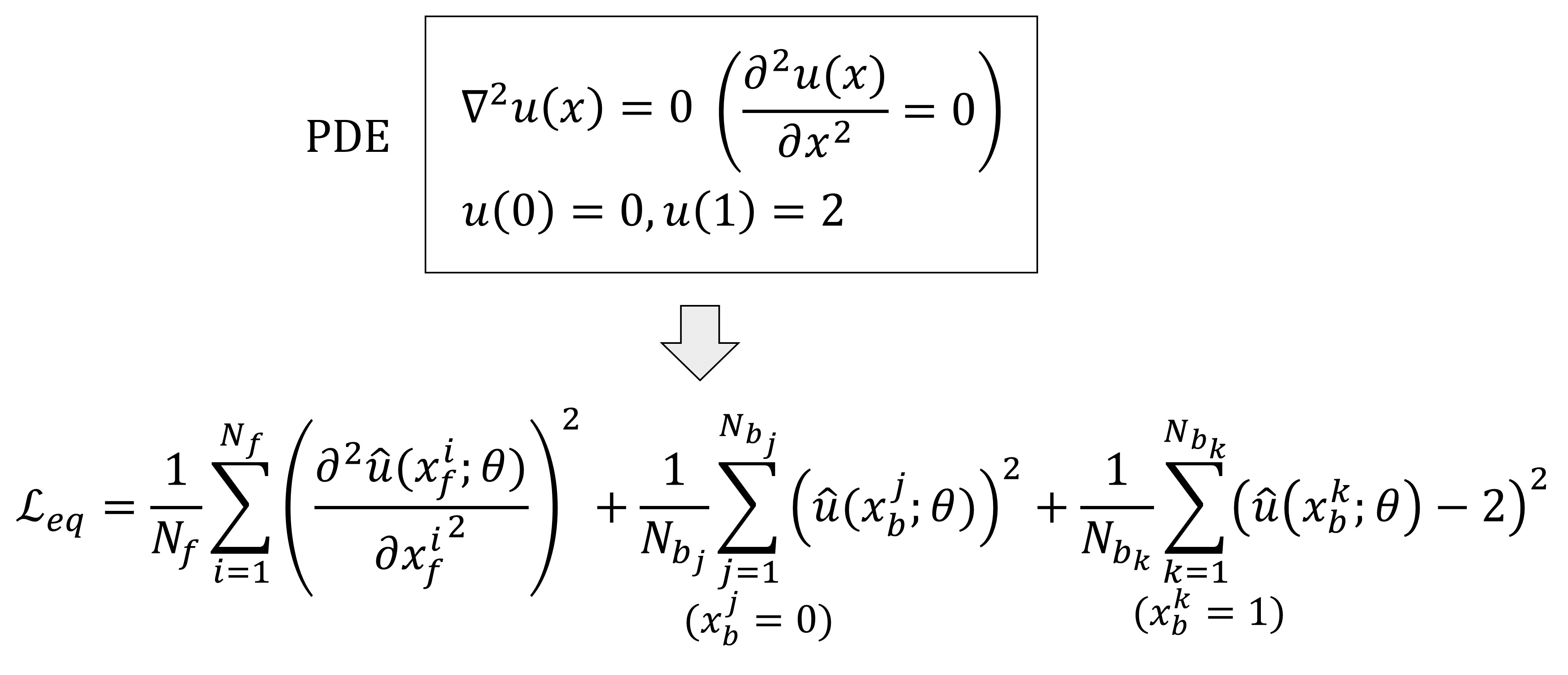}}
  \caption{1D Laplace equation example: We approximate the solution using a neural network, then use automatic differentiation on \(\hat{u}(x; \theta)\) with respect to the input, \(x\), twice to calculate the second-order derivative. \(N_{b_j}\) and \(N_{b_k}\) denotes the number of samples for learning the condition \(u(0)=0\) and \(u(1)=2\).}
  \label{fig:laplace}
  \vspace{-4mm}
\end{figure}

\subsection{Multi-task learning}
Multi-task learning (MTL) tackles multiple tasks concurrently through a learned shared representation. MTL can be viewed from a network architecture perspective and the optimization procedure for joint learning a set of task-specific losses. From the architecture point-of-view, MTL is usually categorized into hard or soft parameter sharing. In hard parameter sharing (e.g. UberNet \cite{kokkinos2017ubernet}), the parameter set is explicitly divided into shared and task-specific layers, leading to a question of which layers the common information should be learned. Thus, soft parameter sharing, such as Cross-stitch \cite{misra2016cross} and Sluice networks \cite{ruder2019latent}, is proposed to adaptively transfer useful information between task-specific networks. However, the scalability to the higher dimension (the number of tasks) of soft parameter sharing is restricted.

More challenges of MTL stem from the optimization process, which balances the impact of each task learning signal. One could deal with the problem by either changing the balancing weights through time or adjusting the gradients of each task-specific loss. Without introducing more hyperparameters, Kendall \textit{et al}. \cite{kendall2018multi} weighted multiple losses by considering the homoscedastic uncertainty (a kind of aleatoric uncertainty) of each task. The weighted loss was derived through maximum likelihood estimation, which was parameterized by the Gaussian noise of each target variable. In our scope, the uncertainty weighting is more preferred than the other methods (e.g. GradNorm \cite{chen2018gradnorm} and DWA \cite{liu2019end}), which either require manual gradient magnitudes balancing or designed and mostly tested with hard parameter sharing architecture. Differently from the uncertainty weighting, Yu \textit{et al}. \cite{DBLP:conf/nips/YuK0LHF20} proposed a so-called PCGrad algorithm to modify the gradients of each task-specific loss so that they were aligned in both direction and magnitude. We choose to compare the effectiveness of uncertainty weighting and PCGrad in terms of the error on unseen data points since both algorithms can be seamlessly applied to existing frameworks of solving PDEs without the need for tuning additional hyperparameters.

\subsection{Adversarial machine learning}
A fascinating property of neural networks, including the state-of-the-art ones, is that they consistently misunderstood adversarial examples, which are easily constructed by adding immeasurably small perturbation on the original inputs (See \cite{DBLP:journals/corr/SzegedyZSBEGF13, DBLP:journals/corr/GoodfellowSS14} for more details). To prevent this vulnerability, \cite{DBLP:journals/corr/SzegedyZSBEGF13, DBLP:journals/corr/GoodfellowSS14} suggests that training with crafted adversarial examples helps regularize the model and reduce the error rate on test sets and the associated adversarial examples, inspiring us to incorporate the transformed samples (See Fig. \ref{fig:transformed_samples}) to the original training data points.

The adversarial examples do not always have to be crafted before training. In a generative adversarial network (GAN), adversarial samples are automatically generated by a generator, which aims to maximize the discriminator loss. In this study, we exploit the GAN heuristics to produce difficult-to-predict samples, which are adaptive through iterations (See Fig. \ref{fig:transformed_samples}(A and B)), for abolishing the weaknesses of the solver network around the relatively high-loss domain regions.

Regarding the inverse problem, in which the aim is to recover the network representation of a PDE from experimental observations, Yang \textit{et al.} \cite{yang2019adversarial} has put forth adversarial inference for joint distribution matching between generated and observed data, obtaining posterior quantification of the uncertainty associated with the predicted solutions. Thus, the purpose of the generated data and the generator loss are clearly distinct from our proposals.

\section{Method}
\subsection{Solving single forward partial differential equation} \label{FPDE}
We consider the general form of PDEs as follows
\begin{equation}
\boldsymbol{\mathcal{N}}[u(t, x); \lambda]=0, x \in \mathbb{R}^D, t \in \mathbb{R}
\end{equation}

\noindent where \(u(t, x)\) denotes the latent solution and \(\boldsymbol{\mathcal{N}}[u(t, x); \lambda]\) is the underlying partial differential equation which is parameterized by \(\lambda\). The typical physics-informed loss function \cite{raissi2019physics} for solving PDE includes both the underlying equation and a boundary condition, given by 
\begin{equation} \label{eq:PDELoss}
\begin{multlined}
\mathcal{L}_{eq}=\frac{1}{N_f}\sum^{N_f}_{i=1}\abs{\boldsymbol{\mathcal{N}}[\hat{u}(t^i_f, x^i_f; \theta); \lambda]}^2+\\
\frac{1}{N_b}\sum^{N_b}_{j=1}\abs{\hat{u}(t^j_b, x^j_b; \theta)-u(t^j_b, x^j_b)}^2
\end{multlined}
\end{equation}

\noindent where \(\hat{u}(t_f, x_f; \theta)\) and \(\hat{u}(t_b, x_b; \theta)\) together denote the predictions of a PDE solver network, parameterized by \(\theta\), for the entire input space. \(N_f\) denotes the number of interior collocation points. \(N_b\) denotes the data points for learning the initial and boundary condition. An elementary PINN loss formulation example of the 1D Laplace is provided in Fig. \ref{fig:laplace}.

\subsection{Auxiliary task generation}
By setting \(\lambda^{aux} = \alpha\lambda\), we are able to acquire an auxiliary task to be learned jointly. In practice, a good choice of \(\alpha\) is achievable via either arbitrary assignment by human or Bayesian optimization \cite{snoek2012practical} to assuredly enforce the similar solution behavior by minimizing the solver network loss, which we shall discuss shortly. After acquiring \( \lambda^{aux}\) for each generated auxiliary task, we turn our focus to how the solver network can learn from multiple physics-informed objectives based on two effective multi-task learning strategies, uncertainty-weighted loss and gradient surgery, namely PCGrad update rule. In order to apply a supervised weighting scheme, we proceed by the assumption that there exists a close neural network approximation \(\hat{u}({t}_f, {x}_f; \theta)\) at the desired accuracy or equivalently \(\norm{\boldsymbol{\mathcal{N}}[\hat{u}({t}_f, {x}_f; \theta); \lambda]}<\epsilon\). According to the Theorem 1. in \cite{van2020optimally}, minimizing Eq. (\ref{eq:PDELoss}), the neural network predictions are close to the true PDE solutions to certain extent. More specifically, for the interior collocation points, we presumably set
\begin{equation} \label{eq:3}
\begin{aligned}
\theta^{*} &= \argmin_\theta \sum^{N_f}_{i=1}\abs{\hat{u}(t^i_f, x^i_f; \theta)-u(t^i_f, x^i_f)}^2\\
&= \argmin_\theta \sum^{N_f}_{i=1}\abs{\boldsymbol{\mathcal{N}}[\hat{u}(t^i_f, x^i_f; \theta); \lambda]}^2.
\end{aligned}
\end{equation}

\subsection{Uncertainty-weighted physics-informed loss function}
As a result of Eq. (\ref{eq:3}), we replace the squared losses in \cite{kendall2018multi} with the typical losses for solving PDEs. We define our uncertainty-weighted physics-informed loss function (Uncert) as follows
\begin{equation} \label{eq:uncert}
\mathcal{L}_{uncert}=\sum^{N_T}_{i=1}(\frac{1}{2\sigma^2_i}\mathcal{L}^i_{eq}+\log\sigma_i) \qquad [Option\:I]
\end{equation}

\noindent where \(N_T\) refers to the number of tasks. Here, \(\sigma_{i}\) is a gradient-based trainable (e.g. by utilizing \(\nabla_{\sigma_{i}}\mathcal{L}_{uncert}\)) parameter which indicates each PDE solution uncertainty. The uncertainty originates under the assumption that, for each PDE \(p(u^{i}(t, x)|\hat{u}^{i}(t, x; \theta))=\mathcal{N}(\hat{u}^{i}(t, x; \theta), \sigma^{2}_{i})\) which fairly encapsulates both the noisy and noise-free cases. Then \(\sigma_{i}\), the scaling factor, is derived to maximize the multi-task Gaussian likelihood, which could be seen as minimizing Bayesian information criterion (BIC) so that the solver network yields the optimal performance for a given solver network parameters \(\theta\), or model complexity.

\subsection{PCGrad: Project conflicting gradients}
Nevertheless, there are alternative approach for learning from multiple objectives. We also apply the backpropagation algorithm with PCGrad updates (Algorithm 1 in \cite{DBLP:conf/nips/YuK0LHF20}) to calculate the modified gradients for the unweighted equation losses. Consequently, we update the model parameters with
\begin{equation} \label{PCGrad}
\begin{gathered}
\delta_{PC}\theta = \sum^{N_T}_{i=1}\textbf{g}^{i}_{PC} \qquad [Option\:II]\\
\{\textbf{g}^{i}_{PC}\} = PCGrad(\{\nabla_{\theta}\mathcal{L}^i_{eq}\})
\end{gathered}
\end{equation}

\noindent where \(\textbf{g}^{i}_{PC}\) refers to the resulting modified gradients by the PCGrad update for the task \(i\). Then, we use the \(\delta_{PC}\theta\) to update the solver network parameters, \(\theta\). For a couple of tasks, PCGrad projects task \(i\)'s gradient onto the normal vector of task \(j\)'s gradient, and vice versa, to deconflict the gradient directions during training.

\begin{figure}
  \centerline{\includegraphics[width=0.5\textwidth]{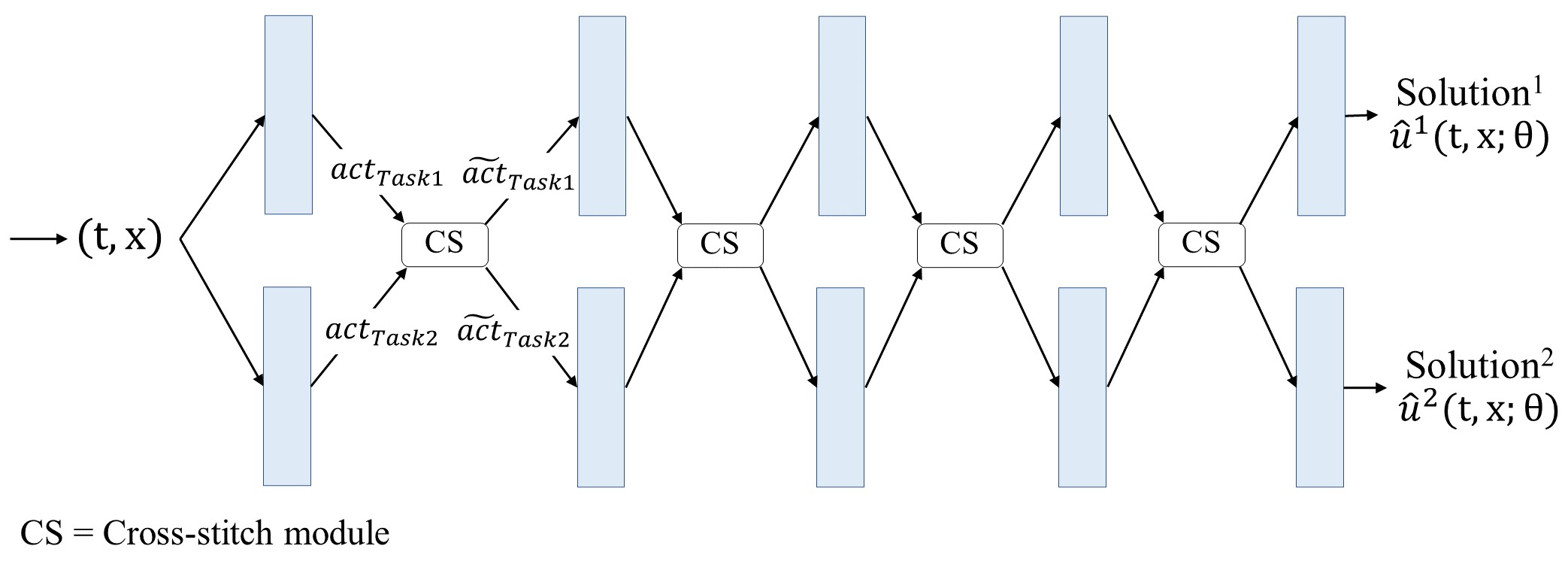}}
  \caption{Overview of the cross-stitch network architecture for the adversarial multi-task training (Algorithm \ref{algo1}).}
  \label{fig:arch}
  \vspace{-4mm}
\end{figure}

\subsection{Multi-task learning architecture}
From the multi-task learning architecture point of view, we leverage two parallel baseline networks with cross-stitch modules to share the activations among
all single-task networks. The overall architecture is depicted in Fig. \ref{fig:arch}. Assume that we are considering two activation maps (\(act_{Task1}, act_{Task2}\)) at a particular layer, which belong to Task1 and Task2 respectively. We note that Task1 is the target PDE and Task2 is the generated auxiliary PDE, A trainable linear transformation of these activation maps is applied, before feeding into the successive layer for each PDE. The transformation can be formalized as
\begin{equation}
\left[ \begin{array}{c} \tilde{act}_{Task1} \\ \tilde{act}_{Task2} \end{array} \right] = \begin{bmatrix} \gamma_{11} & \gamma_{12} \\ \gamma_{21} & \gamma_{22} \end{bmatrix} \times \left[ \begin{array}{c} act_{Task1} \\ act_{Task2} \end{array} \right]
\end{equation}

\noindent where every \(\gamma\) is gradient-based learnable to linearly control how much information to share from Task1 to Task2 and vice versa. With \(\gamma\), cross-stitch modules adaptively retain the low-layer information, for example, \(\gamma_{11}\) determines how much \(act_{Task1}\) influence the higher-layer activations of Task1.

\subsection{Adversarial multi-task training}
Additionally including the high-loss samples lets the solver network focus more on the domain regions that are more challenging to regress for the solutions, for example, the dynamically changing regions of Fokker-Planck solutions shown in Fig. \ref{fig:vis_res}. The generator periodically (every \(F\) iterations, See Algorithm \ref{algo1}.) produces the additional high-loss samples by minimizing the loss function defined as follows

\begin{equation}
\begin{gathered}
\mathcal{L}_{gen}=\frac{1}{N_f}\norm{scale(h(t_f,x_f; \theta_g),lb,ub) - (t_f, x_f)}_2^2-\sum^{N_T}_{j=1}\mathcal{L}^j_{eq}\\
scale(a,lb,ub)=(a-lb)\oslash(ub-lb)\\
\{(t^k, x^k)\}\mysim\{scale(h(t_f^i, x_f^i; \theta_g),lb,ub)\}, \vert\{(t^k, x^k)\}\vert=pN_f\\
\{(t^i, x^i)\}:=concat(\{(t^i,x^i)\}, \{(t^k,x^k)\})
\end{gathered}
\end{equation}

\noindent where \(h(t_f,x_f; \theta_g)\) denotes the transformed samples from the \(N_f\) interior collocation points, \((t_f, x_f)\). The averaged squared \(l_2\)-norm of \(\mathcal{L}_{gen}\) helps the transformed samples to maintain the characteristics of the training distribution. The second term, \(-\sum^{N_T}_{i=1}\mathcal{L}^i_{eq}\), is not averaged, and therefore dominate the overall loss magnitude for inducing the transformed samples to be from the regions that are difficult to regress. We scale the generator's outputs to the PDE domain (bounded from below and above by the vector \(lb\) and \(ub\)) and then uniformly pick a portion, \(p\), of the scaled values to reconstruct \(\{(t^i, x^i)\}\) as the new training set with total \((1+p)N_f+N_b\) samples. The \(\oslash\) and \(\mysim\) notation refer to Hadamard division and uniformly sampling elements from a set. The pseudocode for our adversarial multi-task training is described in Algorithm \ref{algo1}.

\begin{algorithm} \label{algo1}
\caption{Adversarial multi-task training}
\SetAlgoLined
\begin{algorithmic}
\STATE \textbf{Require:} Adversarial training frequency $F$, limit $L$ and iterations $E$, Sampling proportion $p$ and Generator parameters $\theta_{g}$
\FOR{$iter=0$ to $epochs-1$}{
    \IF {$(iter\:mod\:F)=0$ and $iter\leq\floor{\frac{epochs}{L}}$}
        \FOR{$e$ = $0$ to $E-1$}{
            \STATE Freeze the solver network parameters $\theta$
            \STATE Generator's forward pass to obtain $scale(h(t_f,x_f; \theta_g), lb, ub)$
            \STATE Reconstruct $\forall{i}$ $(t^i, x^i)$ as the solver's input
            \STATE Solver's forward pass to obtain $-\sum^{N_T}_{j=1}\mathcal{L}^j_{eq}$
            \STATE Calculate the adversarial loss $\mathcal{L}_{gen}$
            \STATE Backpropagate $\nabla_{\theta_{g}}\mathcal{L}_{gen}$ and update $\theta_{g}$
        }
        \ENDFOR
    \ELSE
            \STATE Freeze the generator parameters $\theta_{g}$
            \STATE Set the latest $\forall{i}$ $(t^i, x^i)$ as the solver's input
            \STATE Solver's forward pass to obtain $\forall{j}$ $\mathcal{L}^j_{eq}$
            \STATE Backpropagate $\nabla_{\theta}\mathcal{L}_{uncert}$ or $\delta_{PC}\theta$ (the options described in Section \ref{FPDE}) and update $\theta$
    \ENDIF
}
\ENDFOR
\STATE \textbf{Return: $\theta^{*}$ and $\theta^{*}_g$}
\end{algorithmic}
\end{algorithm}

\section{Experiments and Results} \label{results}
In this section, we investigate the powerfulness of our methods on a couple of PDE settings, forward partial differential equations and high-dimensional forward-backward stochastic partial differential equations (FBSDE). For each experimental equation of the first setting, trained using the original PINN's loss, 5 fully connected layers with 50 hidden units and tanh activation are considered as the baseline while 4 fully connected layers with 256 hidden units and tanh activation (FBSNN architecture) are selected for experiments of the second case. The sample generator is a single tanh activated hidden layer with 32 units, which exhibits enough capability for obtaining the difficult samples. We always consider an equal number of training and testing samples when comparing our MTL modifications to the typical approach for each PDE.

The performance is quantified using mean absolute error (MAE), mean squared error (MSE) and relative \(l_2\) error. These metrics are defined as follows
\begin{equation}
\begin{gathered}
    MAE = \E[|\hat{u} - u|],\, MSE = \E[(\hat{u} - u)^2]\\
    Relative\:l_2\:error = \frac{\norm{\hat{u} - u}_2}{\norm{u}_2}
\end{gathered}
\end{equation}
\noindent where \(\hat{u}\) is the approximated solution (i.e. the PINN solution), \(u\) is the exact solution, \(\norm{\cdot}_2\) is the \(l_2\)-norm and \(\E\) denotes the expectation over test samples.

\subsection{Forward partial differential equation}
\subsubsection{Burgers' equation}
This equation arises in various areas of applied mathematics, including fluid mechanics, and traffic flow \cite{basdevant1986spectral}. The equation is notoriously hard to solve using traditional numerical methods. We consider the following Burgers' equation with Dirichlet boundary conditions, as being studied in \cite{raissi2019physics}.
\begin{equation}
\begin{gathered}
    u_t + uu_x - (0.01/\pi)u_{xx} = 0,\, x \in [-1, 1],\, t \in [0, 1] \\
    u(0, x) = -\sin(\pi x),\, u(t, -1) = u(t, 1) = 0
\end{gathered}
\end{equation}
Following the training data creation procedure in \cite{raissi2019physics}, a Latin Hypercube Strategy (LHS) \cite{stein1987large} is chosen for sampling 10,000 interior data points (\(N_f\)) and 100 points of the the initial and boundary data (\(N_u\)). The exact solutions for testing are made available by \cite{basdevant1986spectral}. We set \(\lambda=0.01/\pi\) and consider a nearby coefficient of the form \(\lambda^{aux}=\alpha\lambda\). Then, the Bayesian optimization, with \(\mathcal{L}_{uncert}\) in Eq. (\ref{eq:uncert}) as the objective criteria, is applied for searching the best-tuned \(\alpha^{*}\) under the \((0, 1)\) range (for the sake of bounded computational time), obtaining \(\alpha^{*} \approx 0.6\). In the other experiments, we have found that random assignments with minimal tuning are sufficient for training the multi-task networks to outperform the existing approaches (See Section \ref{heuristic} for the effect of \(\alpha\)). Our neural networks are optimized by full-batch Adam \cite{DBLP:journals/corr/KingmaB14} with 0.005 learning rate for 50,000 epochs. For the adversarial setting, we assign \(F = 100\), \(L = 2\), \(E = 10\) and \(p = 0.1\).

\begin{figure*}
  \centerline{\includegraphics[width=0.9\textwidth]{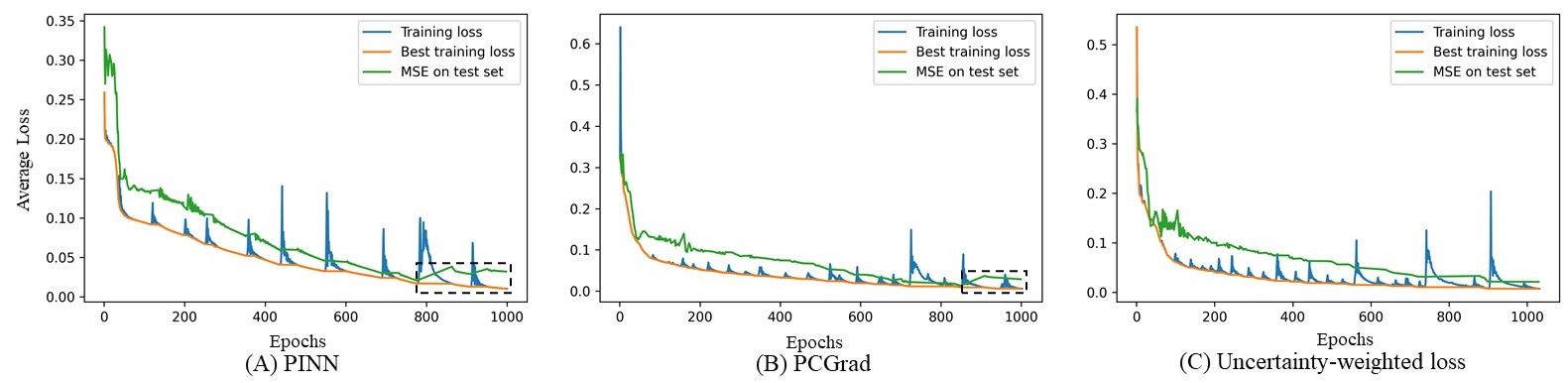}}
  \caption{The average training loss (\(\frac{1}{N_T}\sum^{N_T}_{i=1}\mathcal{L}^i_{eq}\)) plotted with mean squared error (MSE) evaluated on Burgers' equation test set in the case of (A) original PINN (unweighted losses) (B) PCGrad and (C) Uncertainty-weighted loss. The dotted black blocks indicate the overfitting areas. In (B) and (C), where the multi-task learning is employed, the overfitting areas are either narrower (B) or unnoticeable (c). We note that, during training in Section \ref{results}, the network does not have access to any validation or test sets.}
  \label{fig:overfitting}
  \vspace{-4mm}
\end{figure*}

The performance comparison amongst variations of our method, PINN \cite{raissi2019physics} and ResNet \cite{guler2019towards} is listed in Table \ref{table:burger}. Both the multi-task modifications improve the performance by reducing the MAE, MSE and relative \(l_2\) error. We select the best performing strategy, which in this PDE, is the uncertainty-weighting scheme, to undergo our adversarial training, and consequently, the MSE reaches the order of \(10^{-6}\). The adversarial training is found to successfully diminish the losses around the domain regions with high nonlinearity. With the aleatoric uncertainty quantification, Uncert carries out the similar performance against the noisy initial and boundary solutions, signifying our method robustness. Our ablation study of the cross-stitch units, comparing the \nth{4} and \nth{5} row in Table \ref{table:burger} on both the noise-free and noisy cases, shows that the modules are keys for enabling knowledge share between the neural networks, which reduces the overfitting as shown in Fig. \ref{fig:overfitting}.

\setlength{\tabcolsep}{4pt}
\begin{table}
\begin{center}
\caption{Burgers'equation: Performance comparison}
\begin{tabular}{|c|c|c|c|}
\hline
Method & MAE & MSE & Rel. $l_2$ error\\
\hline
PINN \cite{raissi2019physics} & \begin{tabular}{@{}c@{}}$2.3\times10^{-3}$ \\ \textcolor{blue}{$6.7\times10^{-3}$}\end{tabular} & \begin{tabular}{@{}c@{}}$2.1\times10^{-4}$ \\ \textcolor{blue}{$2.1\times10^{-3}$}\end{tabular} & \begin{tabular}{@{}c@{}}$2.4\times10^{-2}$ \\ \textcolor{blue}{$7.4\times10^{-2}$}\end{tabular}\\
ResNet \cite{guler2019towards} & \begin{tabular}{@{}c@{}}$1.8\times10^{-3}$ \\ \textcolor{blue}{$2.8\times10^{-3}$}\end{tabular} & \begin{tabular}{@{}c@{}}$5.0\times10^{-5}$ \\ \textcolor{blue}{$1.7\times10^{-4}$}\end{tabular} & \begin{tabular}{@{}c@{}}$1.1\times10^{-2}$ \\ \textcolor{blue}{$2.1\times10^{-2}$}\end{tabular}\\
\hline
PCGrad w/ CS* & \begin{tabular}{@{}c@{}}$2.3\times10^{-3}$ \\ \textcolor{blue}{$7.1\times10^{-3}$}\end{tabular} & \begin{tabular}{@{}c@{}}$2.7\times10^{-5}$ \\ \textcolor{blue}{$9.3\times10^{-5}$}\end{tabular} & \begin{tabular}{@{}c@{}}$8.4\times10^{-3}$ \\ \textcolor{blue}{$5.7\times10^{-2}$}\end{tabular}\\
Uncert w/o CS* & \begin{tabular}{@{}c@{}}$6.4\times10^{-3}$ \\ \textcolor{blue}{$3.0\times10^{-3}$}\end{tabular} & \begin{tabular}{@{}c@{}}$2.2\times10^{-4}$ \\ \textcolor{blue}{$2.1\times10^{-4}$}\end{tabular} & \begin{tabular}{@{}c@{}}$2.5\times10^{-2}$ \\ \textcolor{blue}{$2.3\times10^{-2}$}\end{tabular}\\
Uncert w/ CS* & \begin{tabular}{@{}c@{}}$1.1\times10^{-3}$ \\ \textcolor{blue}{$1.8\times10^{-3}$}\end{tabular} & \begin{tabular}{@{}c@{}}$2.1\times10^{-5}$ \\ \textcolor{blue}{$1.5\times10^{-5}$}\end{tabular} & \begin{tabular}{@{}c@{}}$7.4\times10^{-3}$ \\ \textcolor{blue}{$6.3\times10^{-3}$}\end{tabular}\\
Uncert w/ CS* + Adv.** & \begin{tabular}{@{}c@{}}$\mathbf{4.1\times10^{-4}}$ \\ \textcolor{blue}{$\mathbf{1.5\times10^{-3}}$}\end{tabular} & \begin{tabular}{@{}c@{}}$\mathbf{1.0\times10^{-6}}$ \\ \textcolor{blue}{$\mathbf{1.0\times10^{-5}}$}\end{tabular} & \begin{tabular}{@{}c@{}}$\mathbf{1.6\times10^{-3}}$ \\ \textcolor{blue}{$\mathbf{5.2\times10^{-3}}$}\end{tabular}\\
\hline
\end{tabular}
\label{table:burger}
\end{center}
\footnotesize\emph{Note:} CS* refers to the cross-stitch module and Adv.** refers to the adversarial multi-task training. These abbreviations are also used in the other tables. Indicated by the blue colour, \(\mathcal{N}(0, 0.01)\) (Gaussian noise) is added to the referenced solutions on the initial and boundary condition, \(u(t_b, x_b)\), for testing the method robustness. The best performance is on boldface.
\vspace{-4mm}
\end{table}
\setlength{\tabcolsep}{1.4pt}

\subsubsection{Poisson equation} The equation is commonly encountered in fluid dynamics. The considering equation is not varying in time, but still, hard to solve analytically and therefore a numerical approach is usually required.
\begin{equation}
\begin{gathered}
    u_{xx} + u_{yy} = f(x, y),\, x \in [0, 1],\, y \in [0, 1]\\
    f(x, y) = -\sin(\pi x)\sin(\pi y)\\
    u(x, 0) = 0,\, u(x, 1) = -\sin(\pi x)\sin(\pi)\\
    u(0, y) = 0,\, u(1, y) = -\sin(\pi)\sin(\pi y)
\end{gathered}
\end{equation}
We sample \(N_f = 8,000\) and \(N_b = 200\) using LHS for training. To build the test set with the shifted distribution, the input space of \((x, y)\) is thoroughly discretized, having \(\Delta x = \Delta y = 0.005\). To generate an auxiliary task, we scale up the \(f(x, y)\), setting \(f^{aux}(x, y)=-2\pi^{2}\sin(\pi x)\sin(\pi y)\) as employed in \cite{dockhorn2019discussion}.  We train our neural networks using full-batch Adam with 0.005 learning rate for 50,000 epochs. For the adversarial setting, we set \(F = 100\), \(L = 5\), \(E = 10\) and \(p = 0.1\).

\begin{figure}
  \centerline{\includegraphics[width=0.4\textwidth]{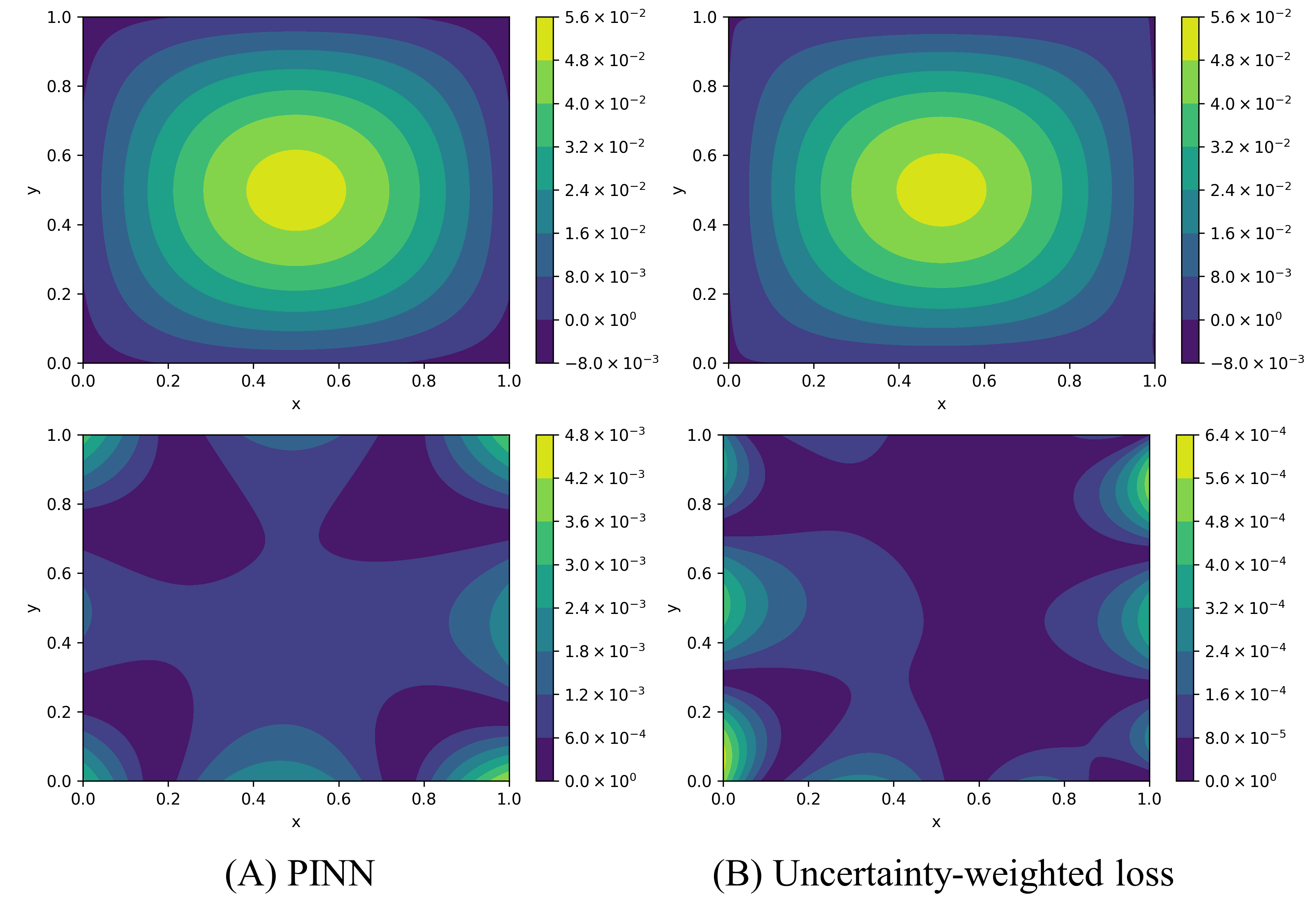}}
  \caption{2D visualization of the Poisson equation's estimated solutions by (A) PINN and (B) our uncertainty weighting based multi-task network enhanced with the adversarial training (the best variant of Table \ref{table:poisson}). At the second row, the color bar indicates the absolute prediction error from the ground truth.}
  \label{fig:poisson_vis}
  \vspace{-4mm}
\end{figure}

The reported results in Table \ref{table:poisson} shows that both the MTL modifications show the greater performance compared to PINN and ResNet, still, there is no significant difference between the strategies; thus the adversarial training is conducted on both strategies. The similar results are found in the PCGrad case whilst the performance boosts are seen in the uncertainty weighting case, contributing to the highest prediction accuracy as shown in Fig. \ref{fig:poisson_vis}. We also found that the cross-stitch units are recommended to make the most out of the designed MTL losses.

\setlength{\tabcolsep}{4pt}
\begin{table}
\begin{center}
\caption{Poisson equation: Performance comparison}
\begin{tabular}{|c|c|c|c|}
\hline
Method & MAE & MSE & Rel. $l_2$ error\\
\hline
PINN \cite{raissi2019physics} & $7.9\times10^{-4}$ & $8.8\times10^{-7}$ & $2.9\times10^{-2}$\\
ResNet \cite{guler2019towards} & $4.7\times10^{-4}$ & $2.7\times10^{-7}$ & $1.7\times10^{-2}$\\
\hline
PCGrad w/ CS & $1.1\times10^{-4}$ & $2.0\times10^{-8}$ & $4.8\times10^{-3}$\\
Uncert w/o CS & $3.9\times10^{-4}$ & $2.1\times10^{-7}$ & $1.5\times10^{-2}$\\
Uncert w/ CS & $1.2\times10^{-4}$ & $2.0\times10^{-8}$ & $4.7\times10^{-3}$\\
PCGrad w/ CS + Adv. & $1.4\times10^{-4}$ & $2.6\times10^{-8}$ & $4.7\times10^{-3}$\\
Uncert w/ CS + Adv. & $\mathbf{9.5\times10^{-5}}$ & $\mathbf{1.6\times10^{-8}}$ & $\mathbf{4.1\times10^{-3}}$\\
\hline
\end{tabular}
\label{table:poisson}
\end{center}
\vspace{-4mm}
\end{table}
\setlength{\tabcolsep}{1.4pt}

\setlength{\tabcolsep}{4pt}
\begin{table}
\begin{center}
\caption{Fokker-Planck equation: Performance comparison}
\begin{tabular}{|c|c|c|c|}
\hline
Method & MAE & MSE & Rel. $l_2$ error\\
\hline
PINN \cite{raissi2019physics} & $2.2\times10^{-3}$ & $8.4\times10^{-6}$ & $9.9\times10^{-3}$\\
ResNet \cite{guler2019towards} & $3.9\times10^{-3}$ & $2.6\times10^{-5}$ & $1.7\times10^{-2}$\\
\hline
PCGrad w/ CS & $1.7\times10^{-3}$ & $4.9\times10^{-6}$ & $7.6\times10^{-3}$\\
Uncert w/o CS & $1.8\times10^{-3}$ & $5.2\times10^{-6}$ & $7.8\times10^{-3}$\\
Uncert w/ CS & $9.0\times10^{-4}$ & $1.4\times10^{-6}$ & $4.0\times10^{-3}$\\
Uncert w/ CS + Adv. & $\mathbf{3.2\times10^{-4}}$ & $\mathbf{1.9\times10^{-7}}$ & $\mathbf{1.5\times10^{-3}}$\\
\hline
\end{tabular}
\label{table:fokker}
\end{center}
\vspace{-4mm}
\end{table}
\setlength{\tabcolsep}{1.4pt}

\subsubsection{1D Fokker-Planck equation} The 1D PDE, from  \cite{xu2020solving}, which describes a snapshot of the probability density functions evolution of stochastic systems, is stated in Eq. (\ref{eq:fp}). 
\begin{equation} \label{eq:fp}
\begin{gathered}
    -[(ax - bx^{3})u(x)]_{x} + \frac{\sigma^{2}}{2}u(x)_{xx} = 0,\, \Delta{x}\sum^{N_{f}}_{i=1}u(x^{i}) = 1\\
    u(-2.2) = u(2.2) = 0,\, (a, b, \sigma, \Delta{x}) = (0.3, 0.5, 0.5, 0.01)\\
\end{gathered}
\end{equation}

As described in \cite{xu2020solving}, we can get the training set and the boundary set with the step length, \(\Delta x = 0.01\), and evaluate the solver network performance on the more fine-grained points constructed with the smaller step length, \(\Delta x = 0.005\). We consider \(a\) to be the PDE parameter \(\lambda\) and set \(\lambda^{aux}=0.5\) (a close value) to simply generate an auxiliary equation. Our neural networks are optimized using full-batch Adam with 0.01 learning rate for 30,000 epochs. For the adversarial setting, we set \(F = 100\), \(L = 3\), \(E = 10\) and \(p = 0.1\).

Based on the performance comparison, which is provided in Table \ref{table:fokker}, we inspect that the neural network, trained with PCGrad, does not outperform PINN. This might be because, in practice, we could fortuitously break the assumptions of PCGrad (See Theorem 2. in \cite{DBLP:conf/nips/YuK0LHF20}), which essentially need to be held for the loss reduction guarantee. We choose the uncertainty-weighting scheme to further experiment with our adversarial training and, in consequence, the performance is enhanced in terms of the reduced MSE from \(10^{-6}\) to \(10^{-7}\).

\begin{figure*}
  \centerline{\includegraphics[width=\textwidth]{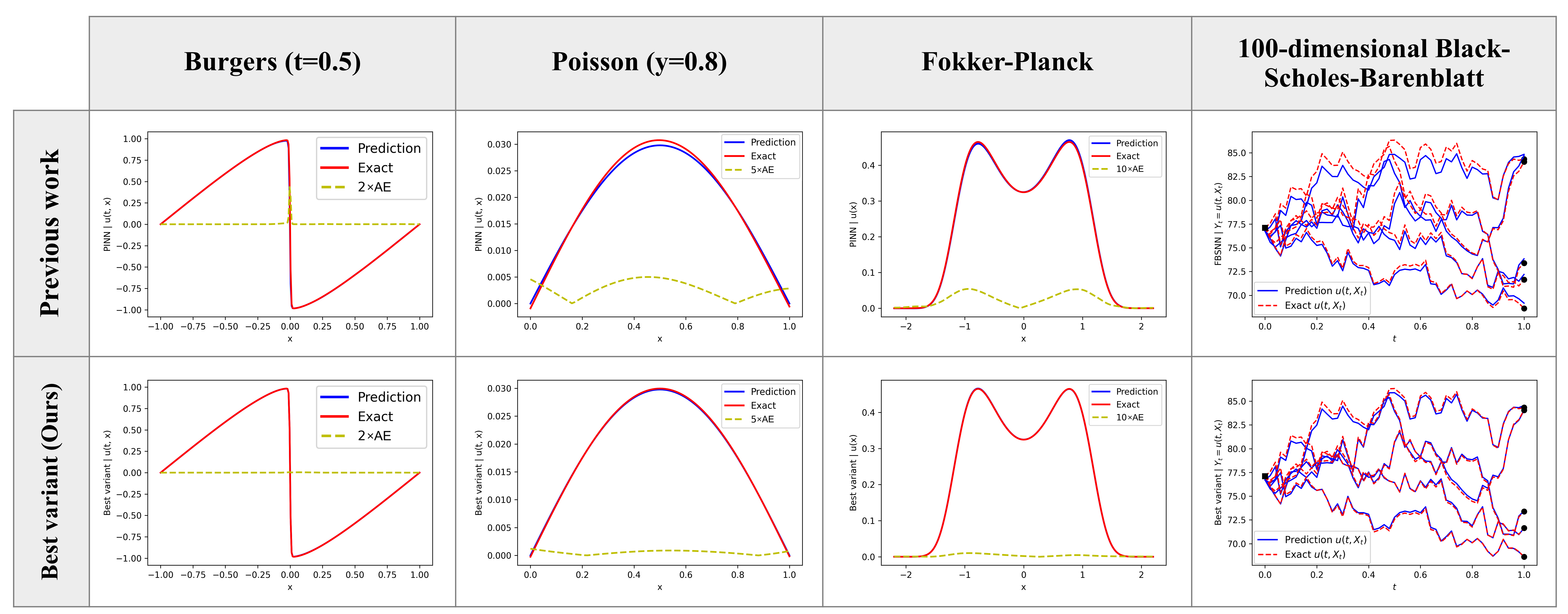}}
  \caption{Comparison of the predicted and exact solutions. The predictions at the first row are reproduced from the previous works (PINN or FBSNN). The predictions at the second row are from the best variant of our proposed method for each PDE. The absolute errors (AEs or dotted yellow lines) around the high-loss or dynamically changing regions are lower when our method is employed.}
  \label{fig:vis_res}
  \vspace{-4mm}
\end{figure*}

\subsection{High-dimensional forward-backward stochastic partial differential equation} \label{high_dimensional_PDEs}
We investigate the effectiveness of applying multi-task learning in the context of solving high-dimensional stochastic differential equations (SDE). Unlike Section \ref{FPDE}, the loss, which is particularly designed to solve SDE, is based on discretizing a SDE using the standard Euler-Maruyama scheme \cite{raissi2018forward}. The designed loss focuses on keeping the relationship between the current state values and the next time-step state values of each stochastic process, \(X_{t}, Y_{t}\) and \(Z_{t}\) rather than estimating the solutions for the entire spatial-temporal grid; therefore the experiment with adversarial training is deducted due to the problem formulation difference and some subtle issues, for instance, Brownian motions at which time steps should be replaced with the transformed ones. This needs further investigation, which is beyond the scope of this paper.

To be more specific we consider coupled forward-backward SDE of the form
\begin{equation}
\begin{gathered}
    dX_t = \mu(t, X_t, Y_t, Z_t)dt + \sigma(t, X_t, Y_t)dW_t, t \in [0, T]\\
    dY_t = \varphi(t, X_t, Y_t, Z_t)dt + Z^{tr}_t\sigma(t, X_t, Y_t)dW_t, t \in [0, T)\\
    X_0 = \xi,\, Y_T = g(X_T),\, Z_t = \nabla Y_t
\end{gathered}
\end{equation}

\noindent where \(W_t\) is a vector-valued Brownian motion, \(X_0 = \xi \in \mathbb{R}^{d}\) (\(X_{t}, Z_{t} \in \mathbb{R}^{d}\) as well) is the initial condition for the forward equation and \(Y_T = g(X_T)\) represents the known terminal condition of the backward equation. \(tr\) is defined to be the transpose operation. In our examples, \(\mu\) and \(\sigma\) are mappings which output in \(\mathbb{R}^{d}\) and \(\mathbb{R}^{d \times d}\) while \(\varphi\) maps to the equal dimension as \(Y_t\), which is a scalar depending on \(t\) and stochastic process \(X_t\). Assumptions on \(\mu\), \(\sigma\) and \(\varphi\) hold based upon \cite{guler2019towards}. For a single trajectory (1 Brownian realization), we could apply the Euler-Maruyama discretization and approximate
\begin{equation}
\begin{multlined}
    X_{n+1} \approx X_{n} + \mu(t_n, X_n, Y_n, Z_n)\Delta t_n + \sigma(t_n, X_n, Y_n)\Delta W_n\\
    Y_{n+1} \approx Y_{n} + \varphi(t_n, X_n, Y_n, Z_n)\Delta t_n +  Z^{tr}_n\sigma(t_n, X_n, Y_n)\Delta W_n\\
\end{multlined}
\end{equation}

\noindent for \(n = 1,2,..,N-1\), where \(\Delta t^n=t^{n+1}-t^n=T/N\) and \(\Delta W_n \sim \mathcal{N}(0, \Delta t_n)\). The approximated error term to be minimized is defined as follows
\begin{equation} \label{FBSNNsLoss}
\begin{multlined}
    \mathcal{L}_{eq}=\sum^{M}_{m=1}\sum^{N-1}_{n=0}|Y^m_{n+1}-\hat{Y}^m_{n+1}|^2 + \sum^{M}_{m=1}|Y^m_N-g(X^m_N)|^2\\
    =\sum^{M}_{m=1}\sum^{N-1}_{n=0}|Y^m_{n+1}-Y^m_{n}-\varphi(t_n, X^m_n, Y^m_n, Z^m_n)\Delta t_n\\
    -(Z^{m}_n)^{tr}\sigma(t_n, X^m_n, Y^m_n)\Delta W^m_n|^2 + \sum^{M}_{m=1}|Y^m_N-g(X^m_N)|^2
\end{multlined}
\end{equation}

\noindent where M and N denote the number of trajectories (batch sizes) and time steps. For every time steps \(n\), we parameterize \(Y_n = u(t_n, X_n; \theta)\) using a single neural network, then \(Z_n\) can be estimated by automatic differentiation of \(u(t_n, X_n; \theta)\) with respect to \(X_n\). The approximated error in Eq. (\ref{FBSNNsLoss}) could be intuitively thought of as the \(\mathcal{L}_{eq}\) defined in Eq. (\ref{eq:PDELoss}) and our proposed multi-task learning modifications are applied in the same manner as described in section \ref{FPDE}. The varied terms for producing an auxiliary SDE, for example, the drift \(\mu\), the volatility \(\sigma\) and even the terminal condition, \(g(X_T)\).  We have tested the effectiveness of our approach for a couple of benchmark problems, the high-dimensional (100D) Black-Scholes-Barenblatt equation and 20D Allen-Cahn equation.

\subsubsection{100D Black-Scholes-Barenblatt equation} One of the most well-known SDE in quantitative finance for modelling the price of various financial derivatives.
\begin{equation}
\begin{gathered}
    dX_{t} = \sigma diag(X_t)dW_t, t \in [0, T]\\
    X_{0} = (1, 0.5, 1, 0.5, ..., 1, 0.5)^{tr} \in \mathbb{R}^{100}\\
    dY_t = r(Y_t - Z^{tr}_tX_t)dt + \sigma Z^{tr}_{t}diag(X_t)dW_t, t \in [0, T)\\
    Y_T = \norm{X_T}^{2}, (T, \sigma, r) = (1, 0.4, 0.05)\\
\end{gathered}
\end{equation}

\noindent To train the solver network, we randomly sample 100 Brownian motions for each epoch, according to the training settings, which are employed in \cite{guler2019towards}. Each motion is discretized into 50 time snapshots. We separately sample another 100 Brownian motions with equal discretized time snapshots for evaluating the network performance after training. We vary the parameter \(\sigma\) which controls the volatility (Gaussian noise) of the stochastic process \(X_t\) from 0.4 to a nearby value, 0.3, to produce the auxiliary SDE. Adam optimizer with 0.001 learning rate is selected for training 20,000 epochs.

The results are listed in Table \ref{table:black} and visualized in the rightmost column of Fig. \ref{fig:vis_res}. PCGrad yields the best performance and enables the predictions to be closer to the exact solutions, including nearby the terminal conditions and the spiky areas where it is harder to regress using the FBSNN or ResNet.

\setlength{\tabcolsep}{4pt}
\begin{table}
\begin{center}
\caption{100D Black-Scholes-Barenblatt equation: Performance comparison}
\begin{tabular}{|c|c|c|c|}
\hline
Method & MAE & MSE & Mean relative error\\
\hline
FBSNN \cite{raissi2018forward} & $4.9\times10^{-1}$ & $3.9\times10^{-1}$ & $6.4\times10^{-3}$\\
ResNet \cite{guler2019towards} & $6.7\times10^{-1}$ & $6.1\times10^{-1}$ & $8.6\times10^{-3}$\\
\hline
PCGrad w/ CS & $\mathbf{2.2\times10^{-1}}$ & $\mathbf{8.3\times10^{-2}}$ & $\mathbf{2.8\times10^{-3}}$\\
Uncert w/ CS & $2.5\times10^{-1}$ & $8.9\times10^{-2}$ & $3.4\times10^{-3}$\\
\hline
\end{tabular}
\label{table:black}
\end{center}
\footnotesize\emph{Note:} The performance results are averaged across 100 unseen Brownian realizations.
\vspace{-4mm}
\end{table}
\setlength{\tabcolsep}{1.4pt}

\subsubsection{20D Allen-Cahn equation}
\begin{equation}
\begin{gathered}
    dX_{t} = dW_t, t \in [0, T]\\
    X_{0} = (0, 0, ..., 0)^{tr} \in \mathbb{R}^{20}\\
    dY_t = (-Y_{t} + Y^3_{t})dt + Z^{tr}_{t}dW_t, t \in [0, T)\\
    Y_T = g(X_T), g(x) = (2 + 0.4\norm{x}^{2})^{-1}, T = 0.3\\
\end{gathered}
\end{equation}

\noindent Following the training and testing scheme employed in \cite{guler2019towards}, we arbitrarily sample 100 Brownian motions for each training epoch. Each motion is discretized into 15 time snapshots. Another 100 Brownian motions with equal discretized time snapshots are sampled for testing the network performance. We generate the auxiliary SDE by slightly changing the terminal condition, setting \(g^{aux}(x) = (2 + 0.3\norm{x}^{2})^{-1}\). Adam optimizer with 0.001 learning rate is selected for training 20,000 epochs.

\begin{figure}
  \centerline{\includegraphics[width=0.5\textwidth]{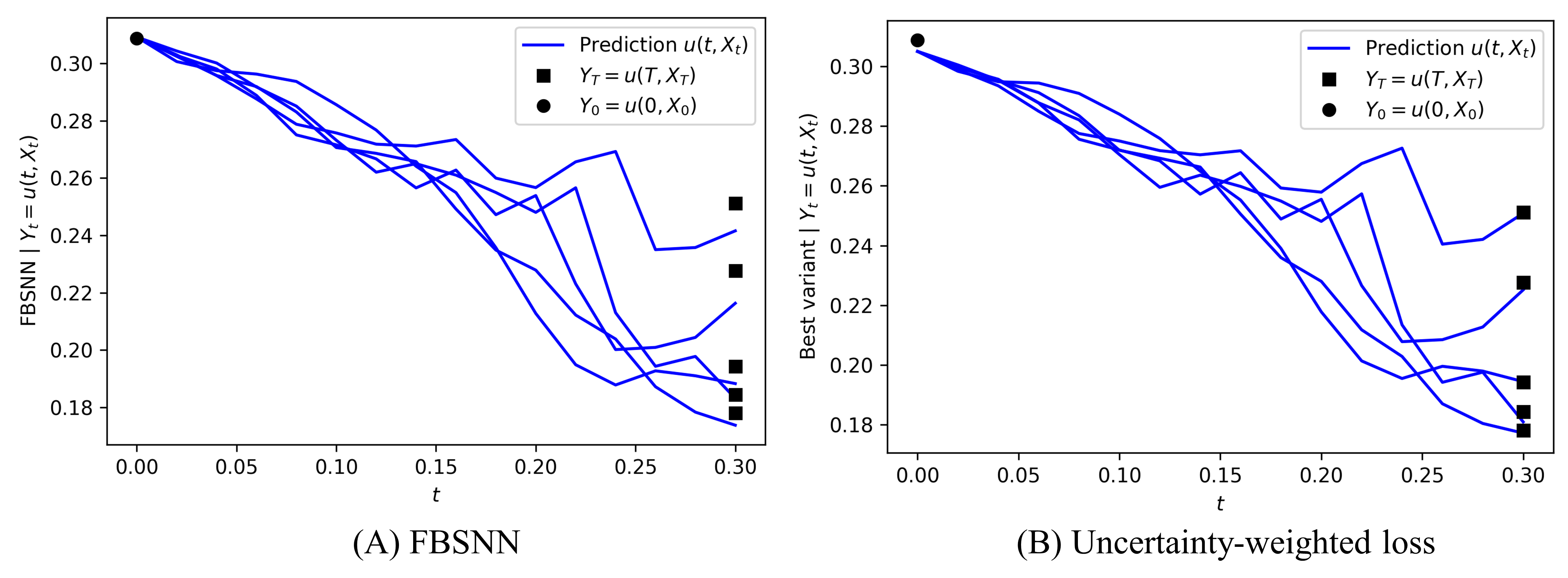}}
  \caption{Predictions by (A) FBSNN and (B) our uncertainty weighting based approach on 5 Brownian motions, which are randomly selected from the test set. More accurate predictions at time \(T=0.3\) are clearly spotted when the uncertainty weighting is applied.}
  \label{fig:allen_visualization}
  \vspace{-4mm}
\end{figure}

We measure the solver network test performance with 100 unseen Brownian motions at the terminal time, \(T=0.3\), where the analytical solution is known. The results listed in Table \ref{table:allen} indicates that the uncertainty-weighted strategy attains the best performance, being competitive to ResNet. This may be because both models perform similar mechanisms for retaining low-level information which helps to suppress the vanishing gradient problem. The visual comparison between FBSNN and the uncertainty weighting approach is shown in Fig. \ref{fig:allen_visualization}.

\setlength{\tabcolsep}{4pt}
\begin{table}
\begin{center}
\caption{20D Allen-Cahn equation: Performance comparison}
\begin{tabular}{|c|c|c|c|}
\hline
Method & MAE & MSE & Mean relative error\\
\hline
FBSNN \cite{raissi2018forward} & $6.7\times10^{-3}$ & $5.1\times10^{-5}$ & $2.8\times10^{-2}$\\
ResNet \cite{guler2019towards} & $3.2\times10^{-3}$ & $1.4\times10^{-5}$ & $1.4\times10^{-3}$\\
\hline
PCGrad w/ CS & $4.1\times10^{-3}$ & $2.3\times10^{-5}$ & $1.7\times10^{-2}$\\
Uncert w/ CS & $\mathbf{2.8\times10^{-3}}$ & $\mathbf{1.2\times10^{-5}}$ & $\mathbf{1.2\times10^{-2}}$\\
\hline
\end{tabular}
\label{table:allen}
\end{center}
\footnotesize\emph{Note:} The performance results are averaged over 100 unseen Brownian realizations at time \(T=0.3\).
\vspace{-4mm}
\end{table}
\setlength{\tabcolsep}{1.4pt}

\subsection{Auxiliary task selection heuristic} \label{heuristic}
\begin{figure}
  \centerline{\includegraphics[width=0.4\textwidth]{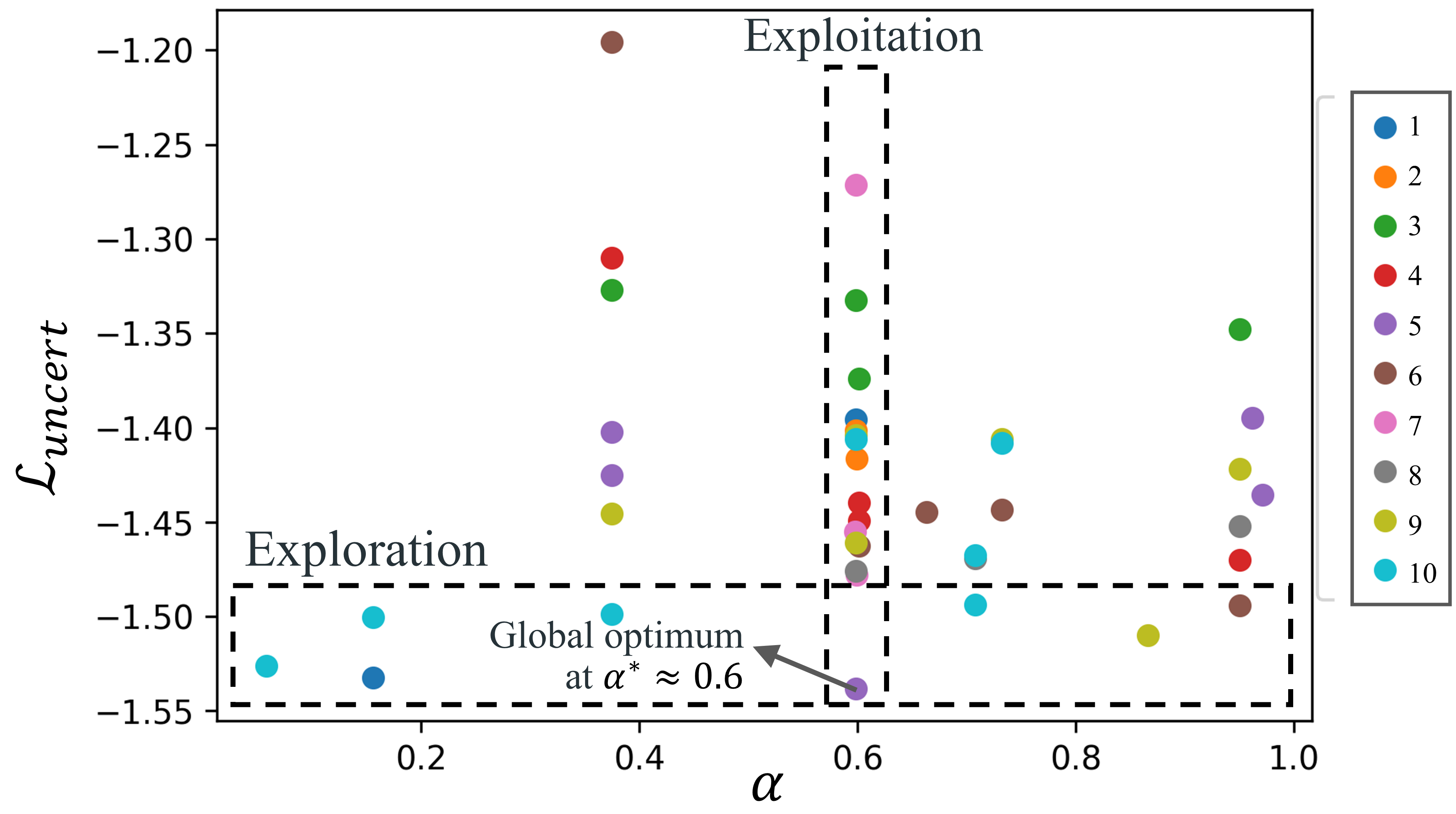}}
  \caption{The results from 10 runs of the Bayesian optimization, excluding the early stopped trials. We frame the exploration box to signify that there are multiple local optimum coefficients, which produce decent performing network checkpoints.
}
  \label{fig:alpha_effect}
  \vspace{-4mm}
\end{figure}

To study the effect of \(\alpha\) in the Burgers' example, we conduct 10 runs of Bayesian optimization with Asynchronous Successive Halving Algorithm (ASHA) \cite{li2018massively} early stopping, which aims to economically search for the optimal alpha 50 trials, minimizing \(\mathcal{L}_{uncert}\) for maximum 200 iterations. The purpose of the multiple runs is to lessen the bias towards a particular random weights initialization and enhance the exploration capacity. The results are shown in Fig \ref{fig:alpha_effect}.

We have found that, with respect to the random weights initialization difference across the multiple Bayesian searches, the various local optimal coefficients are spotted. In the exploitation frame, the network gradually performs better when the coefficient is marginally changed, indicating the loss sensitivity to a group of slightly distinct (not identical) \(\alpha\). Adhering to a local optimal coefficient offers some chances, but does not guarantee obtaining the optimum result in a long run. Since, in practice, the initial neural network weights do not have to be in one particular form, this encourages us that random coefficient assignment is required and might even yield a satisfying trained network as well. To be computationally inexpensive, one should try exploring with the close or random coefficient values to preliminarily obtain the performance. If the network performance does not comply with the requirement, the Bayesian optimization should come to the rescue by increasing the exploitation capability.

\section{Conclusion}
We have introduced the novel approach of applying multi-task learning to the original PINN and FBSNN for producing better-generalized solutions. Compared to the previous works, our method helps reduce the model's test errors on various PDE examples, ranging from low-dimensional settings to high-dimension settings. Both multi-task learning and adversarial training contribute to the enhanced performance of the trained network by incorporating joint representation learning of multiple target PDE instances and additional difficult samples, which helps the network to highly concentrate on the high-loss regions that are more challenging to learn.

As future investigations, our methods are flexible and could be further advanced to tackle (1) chaotic systems, in which two (systems of) equations have similar coefficients, but exhibit completely distinct solutions, (2) parameterized PDEs or (3) a system of PDEs. Finally, together with the development of Physics-guided machine learning such as those studies in \cite{fukui2019physics} and \cite{willard2020integrating}, once the PDE, which governs how the physical system evolves, is effectively solved at the desired accuracy, any learned features embedded in the neural network are easily included in the standard deep learning pipeline for enhancing the network accuracy and robustness.

\section*{Acknowledgment}
This work was supported by JSPS KAKENHI Grant Number JP19K22876.

{\small
\bibliographystyle{ieeetr}
\bibliography{root}
}

\end{document}